\documentclass{article}

\usepackage{graphicx}
\usepackage{color}
\usepackage{subfigure}
\usepackage{tabularx}
\usepackage{amssymb}
\usepackage{latexsym}
\usepackage{amsmath}
\usepackage{algorithm}
\usepackage{relsize}
\usepackage[multiple]{footmisc}
\usepackage{mathtools}
\usepackage{bbold}
\usepackage{multirow}
\usepackage{xfrac}
\usepackage{xspace}
\usepackage{enumitem}
\usepackage{hyperref}

\newcommand{\dv}{\text{d}_{\text{vocab}}}
\newcommand{\dm}{\text{d}_{\text{model}}}
\newcommand{\dhead}{\text{d}_{\text{head}}}
\newcommand{\dmlp}{\text{d}_{\text{mlp}}}
\newcommand{\nh}{\text{n}_{\text{head}}}

\title{Interpreting Affine Recurrence Learning in GPT-style Transformers}

\author{Samarth Bhargav\thanks{Thomas Jefferson High School for Science and Technology, Alexandria, Virginia, USA. Email: 2025sbhargav@tjhsst.edu} \and Alexander Gu\thanks{Thomas Jefferson High School for Science and Technology, Alexandria, Virginia, USA. Email: 2025agu@tjhsst.edu}}

\begin{document}

\maketitle

\begin{abstract}
Understanding the internal mechanisms of GPT-style transformers, particularly their capacity to perform in-context learning (ICL), is critical for advancing AI alignment and interpretability. In-context learning allows transformers to generalize during inference without modifying their weights, yet the precise operations driving this capability remain largely opaque. This paper presents an investigation into the mechanistic interpretability of these transformers, focusing specifically on their ability to learn and predict affine recurrences as an ICL task. To address this, we trained a custom three-layer transformer to predict affine recurrences and analyzed the model's internal operations using both empirical and theoretical approaches. Our findings reveal that the model forms an initial estimate of the target sequence using a copying mechanism in the zeroth layer, which is subsequently refined through negative similarity heads in the second layer. These insights contribute to a deeper understanding of transformer behaviors in recursive tasks and offer potential avenues for improving AI alignment through mechanistic interpretability. Finally, we discuss the implications of our results for future work, including extensions to higher-dimensional recurrences and the exploration of polynomial sequences. 
\end{abstract}

\section{Introduction}

Over the past five years, artificial intelligence (AI) has increasingly permeated daily life, with large language models (LLMs) such as GPT-3 and GPT-4 playing a prominent role. As of 2024, over 180 million users regularly interact with ChatGPT for tasks ranging from writing emails to debugging code and explaining complex concepts \cite{ruby2023chatgpt}. However, as these models demonstrate ever more sophisticated capabilities, our understanding of their internal workings has become increasingly obscured \cite{zhao2023surveylargelanguagemodels}.

The field of AI alignment has raised concerns regarding the rapid development of AI, particularly in the absence of comprehensive interpretability. Without clear insights into how these models function, there is no guarantee that AI systems will consistently align with human morals and values. The potential risks associated with granting autonomy to these models are substantial, with 36\% of AI researchers predicting that artificial general intelligence could lead to catastrophic outcomes on a "nuclear-level" scale \cite{stanford2024aiindex}. Furthermore, over half of AI researchers express grave concern about scenarios involving misinformation and public manipulation \cite{grace2024thousandsaiauthorsfuture}. In light of these concerns, there is an urgent need to develop methods for interpreting and understanding AI.

Mechanistic interpretability is an emerging field that seeks to address this challenge through both empirical and theoretical analyses of neural network weights and biases. A significant portion of mechanistic interpretability research has focused on transformer architectures, such as those employed by ChatGPT, due to their exceptional performance in text generation tasks \cite{li2021pretrainedlanguagemodelstext}.

\section{Background}

For the purposes of this analysis, we adopt the conventions outlined in Appendix A.1. Readers unfamiliar with these concepts are encouraged to consult the appendix for a comprehensive mathematical framework for understanding transformers.

In this section, we explore the capabilities of autoregressive decoder-only transformers, specifically in-context learning (ICL). We also introduce the specific task of mechanistically interpreting affine recurrencess.

\subsection{In-Context Learning}

In-context learning (ICL) refers to the ability of transformers to learn new sequences during inference without adjusting their weights and biases. Formally, given an input sequence $\vec{v}_1, \vec{v}_2, \ldots, \vec{v}_n$, ICL is the ability of a transformer to learn a function $f(\vec{v}_1, \vec{v}_2, \ldots, \vec{v}_n)$ without modifying its internal parameters.

Studies have demonstrated that transformers are capable of performing ICL across a wide range of function classes, including linear regressions, sparse linear functions, two-layer neural networks, and decision trees \cite{garg2023transformerslearnincontextcase}. ICL is foundational to the power of transformers, as exemplified by models like ChatGPT, which must "learn" the intent of a prompt at inference time in order to generate a response.

Despite its centrality to transformer functionality, the mechanisms underlying ICL remain poorly understood. Prior research has reverse-engineered circuits within transformers that facilitate ICL, identifying subsets of attention heads and multi-layered perceptrons (MLPs) that work together to complete specific tasks. One of the most prominent examples of an ICL circuit is the induction head, which is hypothesized to be a fundamental component enabling transformers to generalize \cite{olsson2022incontextlearninginductionheads}.

A concrete example of ICL can be seen in the indirect object identification circuit \cite{wang2022interpretabilitywildcircuitindirect}, a 26-node circuit found in GPT-2 small that successfully predicts the next word in sentences like the following:

\begin{center} 
[Subject 1] and [Subject 2] [verb] [noun]. [Subject 1/2] [verb] \_\_\_\_
\end{center}

For example:

\begin{center}
    Mary and Bob went to the store. Mary gave \_\_\_\_
\end{center}

The correct completion in this case is "Bob." As this example illustrates, the complexity of circuits increases with the complexity of the task, complicating mechanistic interpretation.

\subsection{Affine Recurrences}

In this study, we focus on the task of predicting affine recurrences. An affine recurrence is defined by the following equation:

\[ \vec{a}_n = \begin{cases} 
      \vec{a}_0 & n = 0 \\
      c\vec{a}_{n-1} + \vec{d} & n \geq 1 \\
   \end{cases}
\]

The specific task we address involves training a transformer to auto-regressively predict the sequence $\vec{a}_1, \vec{a}_2, \ldots, \vec{a}_n$ from the inputs $\vec{a}_0, \vec{a}_1, \ldots, \vec{a}_{n-1}$. The challenge lies in making these predictions without knowledge of the values of $c$ or $\vec{d}$.

\subsection{Related Works}

The study of mechanistic interpretability has gained significant traction in recent years, driven by the need to understand how large language models (LLMs) like GPT-3 and GPT-4 perform complex tasks without explicit supervision. The concept of in-context learning (ICL), wherein transformers adapt to new tasks during inference without updating their weights, is a cornerstone of this research.

Garg et al. \cite{garg2023transformerslearnincontextcase} explored the capacity of transformers to perform in-context learning across various simple function classes, including linear regressions, decision trees, and sparse linear functions. Their findings laid the groundwork for understanding how transformers can generalize during inference, demonstrating that transformers can learn tasks of significant complexity purely through ICL mechanisms.

In parallel, research by Olsson et al. \cite{olsson2022incontextlearninginductionheads} focused on reverse-engineering the internal circuits of transformers during ICL. They identified specific patterns, such as the induction head circuit, which is hypothesized to play a key role in enabling transformers to generalize by copying tokens from previous contexts. This study provided a critical foundation for the current work by highlighting how transformers might leverage specific heads and layers to perform in-context learning tasks.

Nanda et al. \cite{transformer2021framework} extended these insights by developing a mathematical framework for analyzing the QK (Query-Key) and OV (Output-Value) circuits in transformers. Their work introduced the notion of eigenvalue scores to measure whether certain heads act as copying mechanisms within these circuits, a concept directly relevant to understanding the behavior of our model in predicting affine recurrences.

While there has been considerable research on interpreting transformers in the context of text generation and ICL, studies specifically targeting affine recurrences and other recursive sequences within numerical tasks are sparse. The closest related work is the analysis of Fibonacci sequences in transformers, identified as one of the open problems in mechanistic interpretability by Nanda \cite{interpretability2019concrete}. Our work builds on these foundations, exploring a more complex class of recurrences and extending the analysis to multiple layers and attention heads within a transformer.

Finally, the Automated Circuit Discovery (ACDC) approach, as outlined by Conmy et al. \cite{conmy2023automatedcircuitdiscoverymechanistic}, represents an emerging direction in the field. ACDC aims to automate the identification and analysis of circuits within transformers, which could be instrumental in scaling the mechanistic interpretability efforts to more complex models and tasks. Our research aligns with this vision, contributing empirical insights that could help refine such automated techniques in the future.

In summary, this work positions itself at the intersection of mechanistic interpretability and recursive sequence prediction in transformers. By focusing on affine recurrences, we address an underexplored area within the existing literature, offering new insights into how transformers can learn and refine complex numerical tasks in-context.

\section{Experimental Setup}

\subsection{Data Sampling}

The primary concern during data generation is that the linear coefficient in the affine recurrence, denoted by $c$, could potentially be very large, leading to exponential growth in the sequence. This exponential growth presents a challenge for stochastic gradient descent (SGD), the optimization method typically used to train transformers. Large gradients, arising from the partial derivatives of the error function with respect to model parameters, can destabilize the training process.

To mitigate this issue, we bounded the value of $c$. Even with this bounding, a relatively small number of terms (e.g., 10) can result in the sequence growing by a factor of 1,000. To ensure the model trains properly under these conditions, we employed the following data normalization steps:

\begin{itemize}[itemsep=0pt, parsep=0pt, topsep=0pt, partopsep=0pt]
    \item Calculate the first $n$ terms of the recurrence, initializing $\vec{a}_0$, $c$, and $\vec{d}$ with uniformly random numbers from the range $[-2, 2]$.
    \item Compute the maximum norm of the vectors $\vec{a}_0, \vec{a}_1, \ldots, \vec{a_{n-1}}$, denoted by $m$.
    \item Divide all vectors by $\frac{m}{\text{rand}(1,2)}$, where $\text{rand}(x,y)$ returns a random number in the range $[x,y]$. Similarly, divide $\vec{d}$ by this same quantity.
    \item This process defines a new recurrence that is normalized so that the maximum element has a norm between $1$ and $2$.
\end{itemize}

 This normalization procedure ensures that the model does not encounter excessively large gradients, thereby smoothing the training process. Importantly, the approach preserves the generality of affine recurrences, as any such sequence can be normalized before being input into the model.

An important insight from our data generation process is that each principal axis of the vectors $\vec{a}_n$ contains its own affine recurrence. Consequently, $\dim(\vec{a}_n)$ distinct affine recurrences are running simultaneously (assuming they share the same linear term), and each can be analyzed independently.

\begin{figure}[t]
    \centering
    \includegraphics[width=0.5\linewidth]{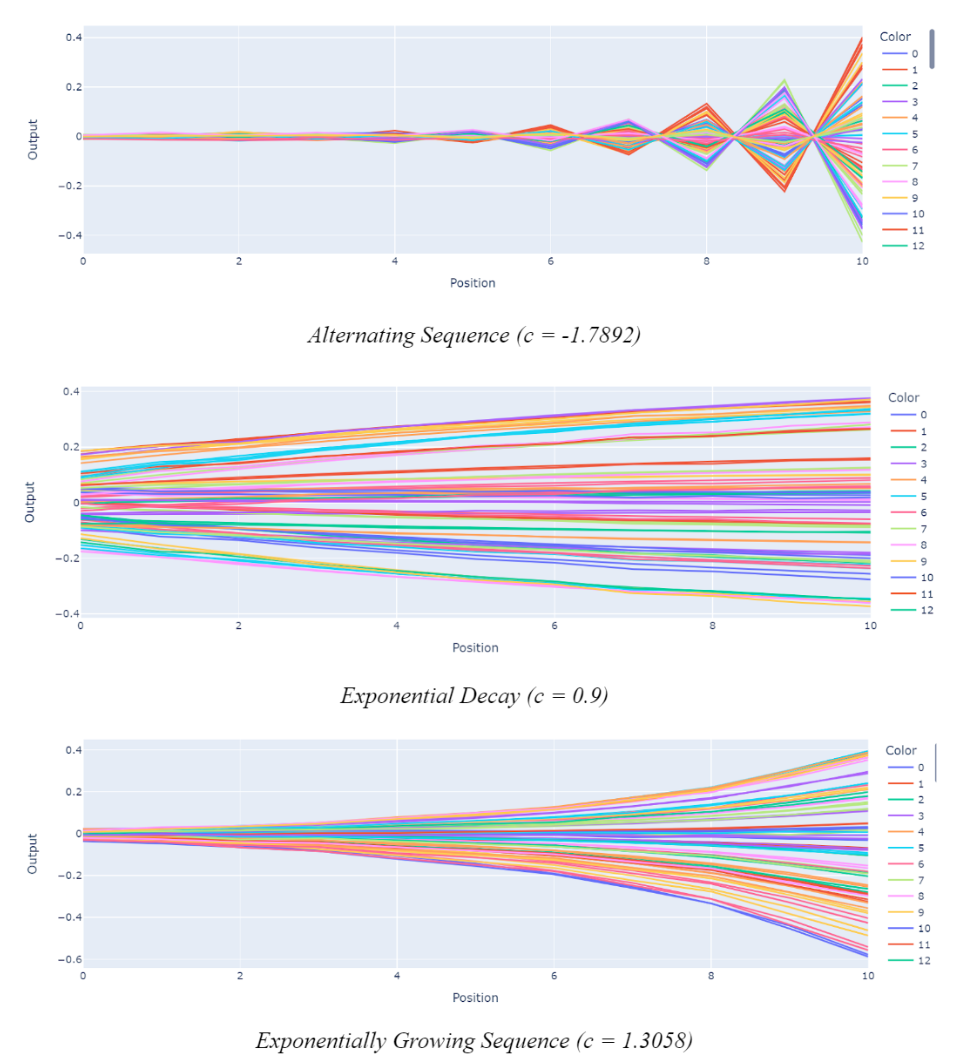}
    \caption{Types of Affine Recurrences.}
    \label{fig:sequencetypes}
\end{figure}

To illustrate the diversity of the sequences generated, we provide examples
of three distinct types of sequences shown in Figure ~\ref{fig:sequencetypes}. The first example involves a negative $c$, resulting in an oscillating sequence. The second example features a positive $c$ with $0 < c < 1$, leading to exponential decay. The final example involves a positive $c > 1$, resulting in exponential growth. Each color in Figure \ref{fig:sequencetypes} represents a different principal axis of the $\vec{a}_n$ vectors.

Finally, we constrained the sequence length to be between 3 and 14. The upper bound was imposed to prevent the initial elements from approaching zero, while the lower bound was enforced for reasons discussed in the model architecture section.

\subsection{Model Architecture}

We employed a custom three-layer transformer architecture with eight attention heads per layer. The architecture follows the framework detailed in Appendix A.1. The specific parameters of the model are as follows:

\begin{align*}
    \dv &= \dim(\vec{a}_n) = 40\\
    \dm &= 128\\
    \dhead &= 64\\
    \dmlp &= 3072\\
    \nh &= 8\\
    \text{n}_{\text{layers}} &= 3\\
    \text{n}_{\text{ctx}} &= 32
\end{align*}

Notably, TransformerLens, a well-regarded library for the mechanistic interpretation of GPT-style models, does not natively support models that bypass tokenization. To accommodate our needs, we created a custom fork of TransformerLens that supports direct vector inputs. The link to this modified version is provided in the appendix.

The model directly receives the vectors $\vec{a}_0, \vec{a}_1, \ldots, \vec{a}_{n-1}$ as input and is expected to output the vectors $\vec{a}_1, \vec{a}_2, \ldots, \vec{a}_n$. We employed mean-squared error (MSE) as the loss function for training.

Notably, we did not apply the MSE loss uniformly across all vectors. The rationale is that the first vector the model could predict correctly is $\vec{a}_3$. To illustrate this, consider the following relationships:

\begin{align*}
    \vec{a}_1 &= c\vec{a}_0 + \vec{d}\\
    \vec{a}_2 &= c\vec{a}_1 + \vec{d}\\
\end{align*}

It is evident that $c$ and $\vec{d}$ can be extracted from this system of equations by substituting the known values for $\vec{a}_0, \vec{a}_1, \vec{a}_2$. However, predicting $\vec{a}_2$ solely based on $\vec{a}_0$ and $\vec{a}_1$ is impossible. Empirically, we observed that applying MSE to all vector pairs yielded worse results than starting from the prediction of the vector following $\vec{a}_2$. This observation also motivated the minimum recurrence size of 3 enforced in the previous section.

The model was trained for 100,000 steps using a batch size of 16. We employed the AdamW optimizer with a weight decay of 0.01 and a learning rate of 0.0001. For each batch, the sequence length was randomly chosen to be between 3 and 14. Training was conducted on an NVIDIA T4-GPU using Google Colab.

\section{Discussion}

\subsection{Model Results}

\begin{figure}[H]
    \centering
    \includegraphics[width=0.9\linewidth]{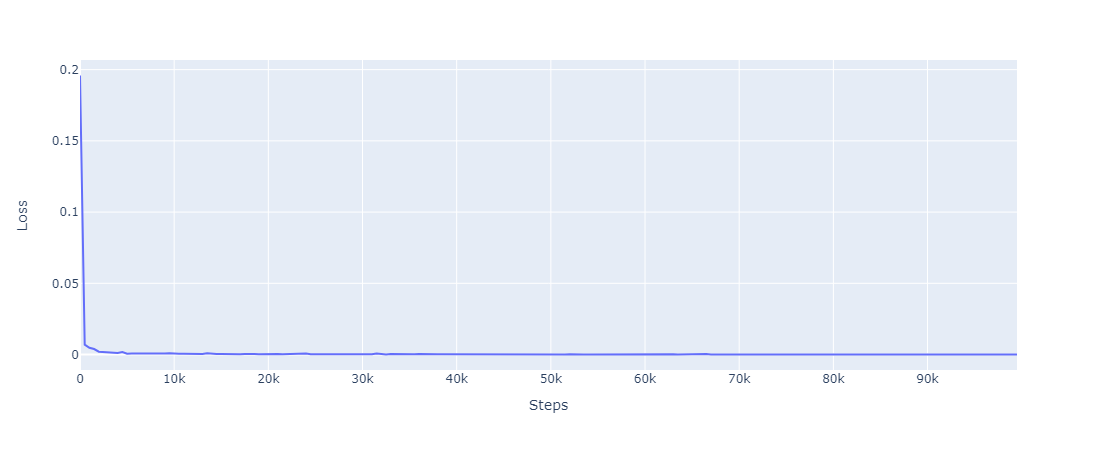}
    \caption{Loss as a Function of Steps.}
    \label{fig:loss_fn}
\end{figure}

The transformer model successfully achieved a mean squared error of 0.0001. Figure \ref{fig:loss_fn} illustrates the model's loss as a function of the number of training steps.

In further analysis of our model, we folded layer normalization weights and value biases to facilitate direct analysis of the $OV$ circuit without the confounding effects of value bias \cite{transformer2021framework}.

\subsection{Attention Patterns}

\begin{figure}[H]
    \centering
    \includegraphics[width=0.75\linewidth]{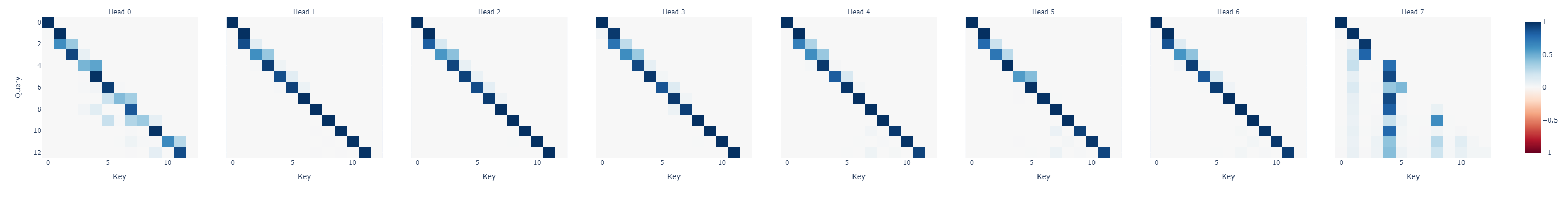}
    \caption{Layer 0 Attention Patterns.}
    \label{fig:L0_attn}
\end{figure}

Attention patterns reflect the weights of the linear combinations of $OV$ vectors that are added to the residual stream vectors. Figure \ref{fig:L0_attn} depicts the attention patterns in the zeroth layer.

In the zeroth layer (indexed from zero), all attention appears to be directed toward the previous vector, except for the last head. This pattern is consistent with what is known in transformer nomenclature as a \textbf{previous token head}.

\begin{figure}[H]
    \centering
    \includegraphics[width=0.75\linewidth]{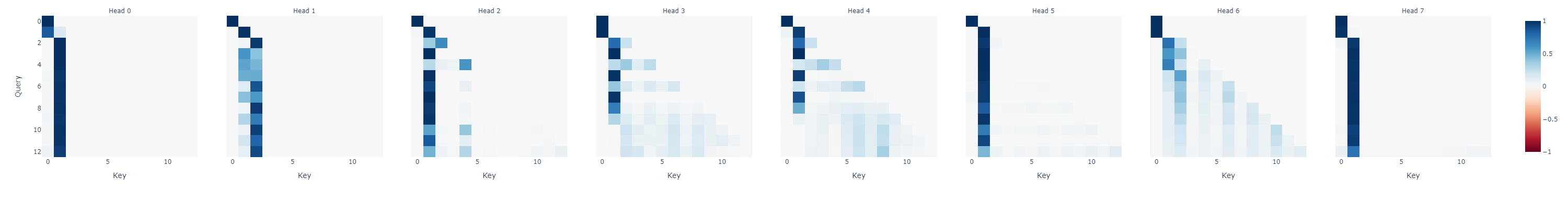}
    \caption{Layer 1 Attention Patterns.}
    \label{fig:L1_attn}
\end{figure}

In the first layer attention patterns shown in Figure \ref{fig:L1_attn}, the majority of attention heads focus specifically on the second vector.

\begin{figure}[H]
    \centering
    \includegraphics[width=0.75\linewidth]{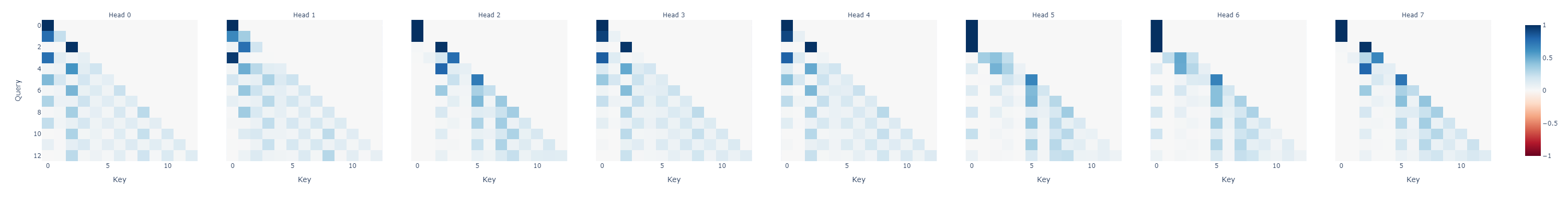}
    \caption{Layer 2 Attention Patterns (Alternating Sequence).}
    \label{fig:L2_attn_alt}
\end{figure}

\begin{figure}[H]
    \centering
    \includegraphics[width=0.75\linewidth]{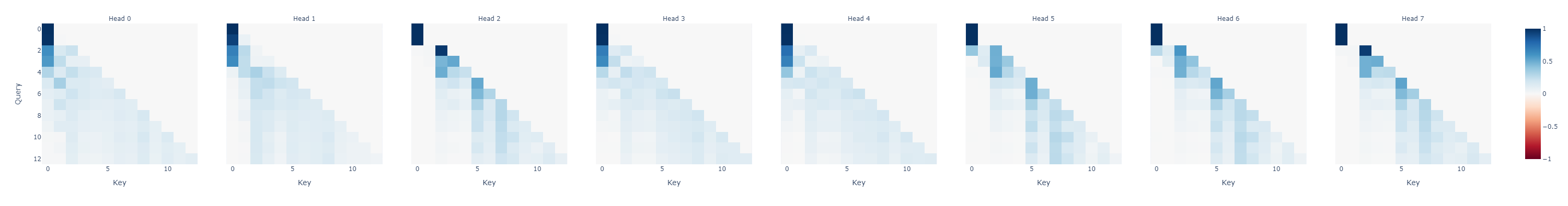}
    \caption{Layer 2 Attention Patterns (Non-Alternating Sequence).}
    \label{fig:L2_attn_nonalt}
\end{figure}

Finally, in the second layer, the attention patterns shown in Figure \ref{fig:L2_attn_alt} exhibit a checkerboard-like structure, reminiscent of the alternating sequence mentioned earlier. Notably, when the affine recurrence is non-alternating, the attention pattern differs significantly, as shown in Figure \ref{fig:L2_attn_nonalt}.

As we will demonstrate later, there is a compelling reason why these checkerboard patterns emerge for alternating affine recurrences but not for non-alternating ones.

\subsection{Direct Logit Attributions}

Direct logit attribution is a key technique for studying how different heads contribute to accurate predictions.

The method involves isolating the output of each attention head and assessing its contribution to the correct prediction of the next vector in the sequence. Specifically, for a given head $H$ that adds the vector $\vec{v'}_m$ to the residual stream at sequence position $m$, we evaluate how much $\vec{v'}_m$ aligns with the target vector $\vec{a}_{m+1}$.

First, we apply the final layer normalization, $\text{ln}_\text{final}$, to $\vec{v'}_m$, followed by multiplication with the unembedding matrix $W_U$ to project $\vec{v'}_m$ back into the unembedding space.

\begin{figure}[H]
    \centering
    \includegraphics[width=0.75\linewidth]{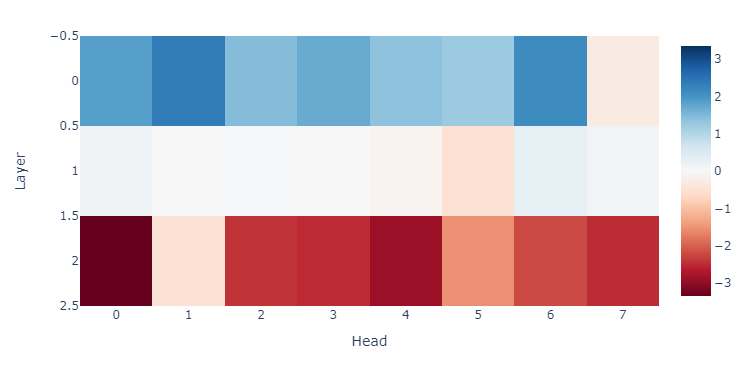}
    \caption{Direct Logit Attribution.}
    \label{fig:DLA_real}
\end{figure}

To quantify the contribution, we compute the dot product between the two vectors $\langle (\text{ln}_{\text{final}} (\vec{v'}_m) W_U, \vec{a}_{m+1} \rangle$. This dot product serves as a measure of similarity between the vectors, indicating how much $\vec{v'}_m$ contributes to the correct prediction. We display these quantities for each head's output as shown in Figure \ref{fig:DLA_real}.

An alternative approach to defining direct logit attribution involves using the dot product with $\vec{a}_{m+1} - \vec{a}_m$, providing a metric for how much the model's output shifts in the "correct" direction from the starting point. In practice, both methods yield nearly identical results.

The direct logit attribution graphs reveal a distinctive pattern: heads in the zeroth layer add information in the correct direction, while heads in the first layer contribute to the orthogonal complement of the embedding space, making their direct effects less apparent. Strikingly, heads in the second layer appear to add information in the opposite direction of the output vector.

From these observations, we hypothesize that the zeroth layer heads overestimate the next vector, and the second layer heads correct this overestimate by adjusting in the opposite direction.

\subsection{Zeroth Layer OV Circuits}

While the attention pattern in the zeroth layer indicates a strong focus on the previous token, we still lack detailed information about the $OV$ circuits in this layer.

\begin{figure}[H]
    \centering
    \includegraphics[width=0.75\linewidth, trim=0 0 0 0, clip]{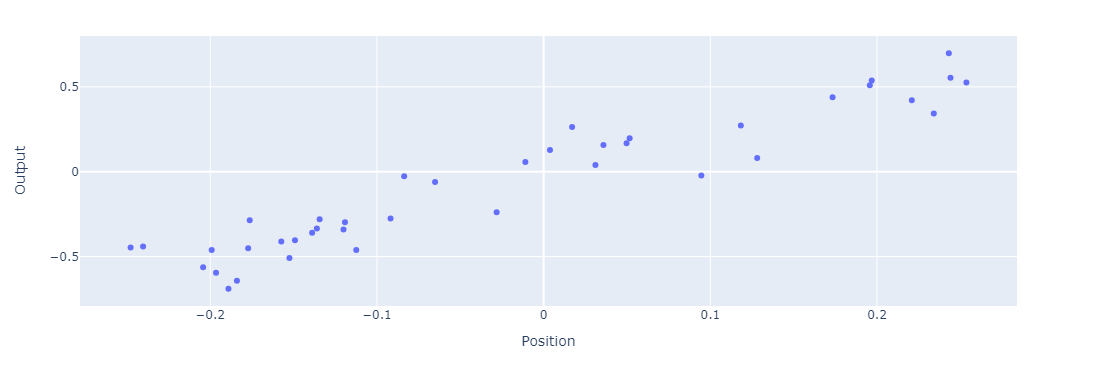}
    \caption{OV Values of Residual Stream Vectors in Layer 0.}
    \label{fig:OVofResid}
\end{figure}

To gain insight, we examined the result of applying $OV$ to each residual stream vector $\vec{x}$. Specifically, we plotted $(\vec{x}_d, (OV\vec{x})_d)$ for each dimension $0 \leq d < \dm$, as shown in Figure \ref{fig:OVofResid}.

Surprisingly, the points form a nearly linear relationship, with an $R^2$ value of 0.83—a significant result. Over multiple residual stream vectors, the y-intercept remains close to zero, while the slope varies between 2.1 and 2.5. This suggests that the zeroth layer approximately performs the operation $\vec{v}_n \rightarrow \vec{v}_n + \alpha \vec{v}_{n-1}$, where $\alpha \in (2.1, 2.5)$.

\begin{figure}[H]
    \centering
    \includegraphics[width=0.75\linewidth]{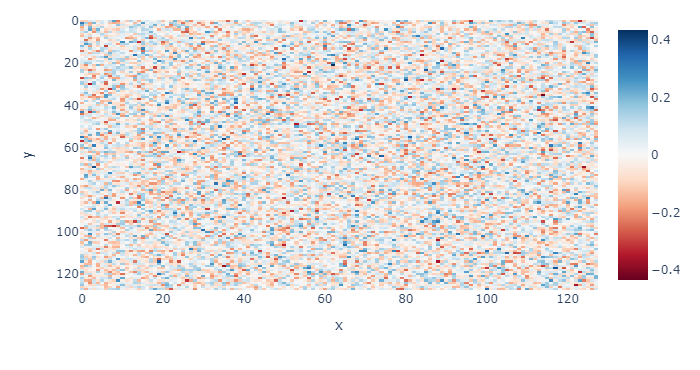}
    \caption{Sum of OV Circuits over all Heads in 0th Layer.}
    \label{fig:sumoverOV}
\end{figure}

A potential concern is that, visually as shown in Figure \ref{fig:sumoverOV}, the zeroth layer heads do not resemble copying heads, even though this analysis suggests they might be.

\begin{figure}[h!]
    \centering
    \includegraphics[width=0.75\linewidth]{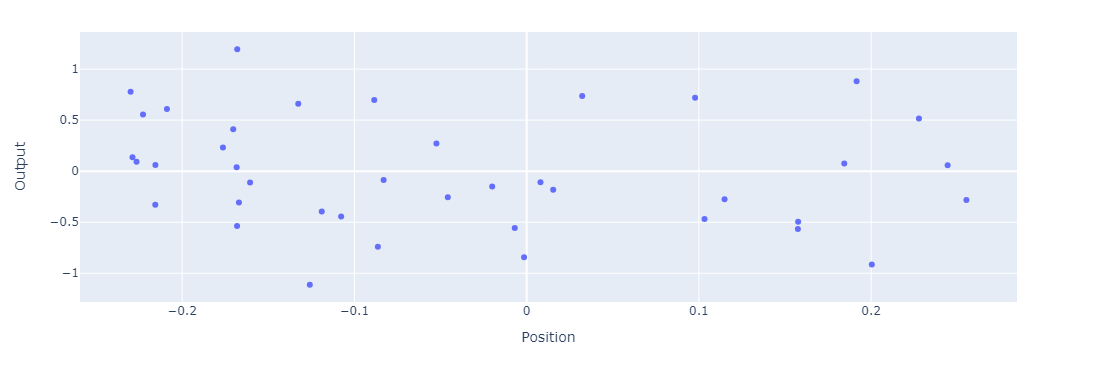}
    \caption{OV Circuit Applied to Random Vector}
    \label{fig:randomov}
\end{figure}

We hypothesize that the $OV$ circuits act as copying heads within the embedding space while simultaneously cross-pollinating information into the orthogonal complement of the embedding space. We tested this hypothesis in two ways: first, by inputting a random, non-residual stream vector into the $OV$ matrix and comparing the results. The results are shown in Figure \ref{fig:randomov}.

While there is still a positive correlation, the $R^2$ values for random vectors averaged 0.598, compared to 0.832 for residual stream vectors. This indicates that the $OV$ circuit acts more like a copying head for residual stream vectors than for random vectors, suggesting that some information is added to the orthogonal complement of the embedding space.

The second approach involved computing the norm of the part of $OV \vec{x}$, where $\vec{x}$ is a residual stream vector, that lies outside the span of the embedding space. This provides a metric for how much information is being added to the orthogonal complement of the embedding space.

To compute this, we performed a singular value decomposition (SVD) of the embedding matrix $W_E = U \Sigma V^T$, where $U$ and $V$ are $\dv \times \dv$ and $\dm \times \dm$ orthogonal matrices, respectively, and $\Sigma$ is a $\dv \times \dm$ diagonal matrix. We then computed the quantity by subtracting the part of $OV \vec{x}$ that lies along each row in $V^T$. Since the rows of $V^T$ are orthogonal, taking the norm of the subtraction yields the portion of $OV\vec{x}$ that lies in the orthogonal complement of the embedding space. Finally, we took the ratio of this norm to the original norm of $OV\vec{x}$ to determine the ratio of information being added in the orthogonal complement.

We found that this quantity was 84.15\%, indicating that, on average, 84.15\% of the information moved by the $OV$ circuits in layer 0 resides in the orthogonal complement of the embedding space. This explains why the $OV$ circuits do not visually resemble identity matrices.

\subsection{First Layer}

Given that the direct logit attribution scores of all heads in the first layer are near zero, it is natural to question whether these heads are useful.

A common technique for assessing the utility of specific heads is ablation, which involves removing a particular head to analyze its impact on accuracy.

There are two primary types of ablation: zero ablation and mean ablation. Zero ablation replaces the output of a head with the zero vector, while mean ablation replaces it with the mean output of that head over a distribution of inputs. Intuitively, zero ablation may seem preferable; however, as Neel Nanda argues in \cite{transformer2021framework}, zero ablation can be unprincipled if the head primarily adds a "bias" term (e.g., consistently adding a value of 100 to the residual stream). In such cases, zero ablation could significantly reduce model performance, even if the head's output does not depend on the input.

We implemented mean ablation using a population dataset of 1,000 samples and observed a new mean squared error of 0.0002. This is very close to the original mean squared error of 0.0001, leading us to conclude that the first layer heads' outputs do not depend significantly on their inputs and are therefore comparatively less useful.

\begin{figure}[H]
    \centering
    \includegraphics[width=0.75\linewidth]{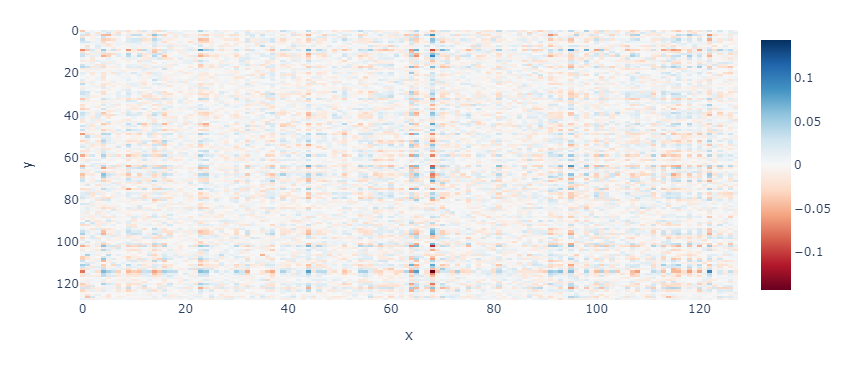}
    \caption{OV Circuit of Layer 1 Head 0}
    \label{fig:L1OV}
\end{figure}

Another perspective on this issue can be gained by directly examining the $OV$ circuits in the first layer. As shown in Figure \ref{fig:L1OV}, the $OV$ matrix appears sparse and exhibits low effective rank, supporting the hypothesis that the $OV$ matrix only adds information in a single feature direction, effectively acting as a "bias" term.

\subsection{Second Layer QK/OV Circuits}

\begin{figure}[H]
    \centering
    \includegraphics[width=0.75\linewidth]{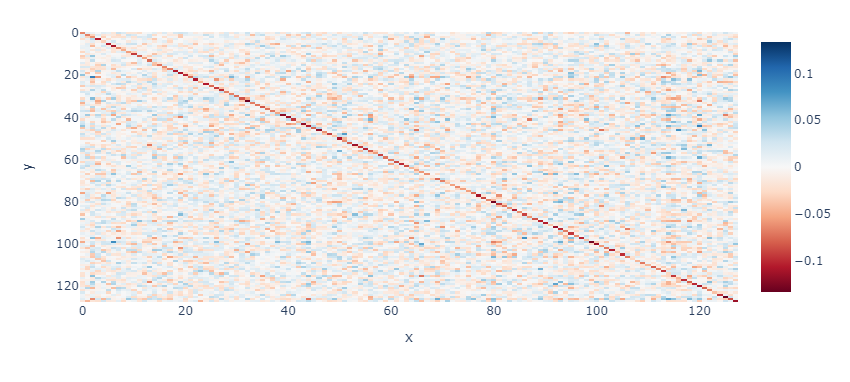}
    \caption{OV Circuit of Layer 2 Head 0}
    \label{fig:L2OV}
\end{figure}

\begin{figure}[H]
    \centering
    \includegraphics[width=0.75\linewidth]{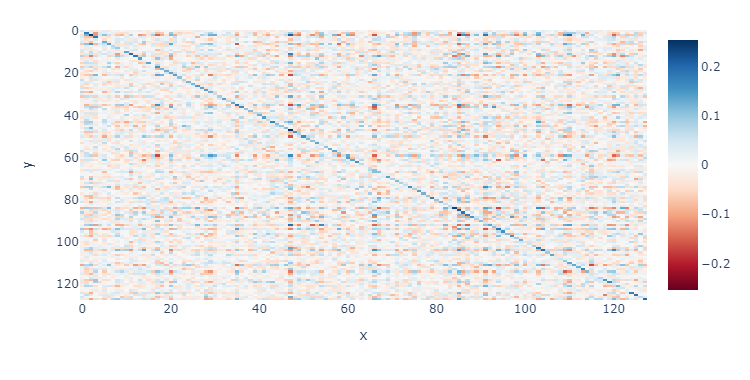}
    \caption{QK Circuit of Layer 2 Head 0}
    \label{fig:L2QK}
\end{figure}

When plotting the $OV$ and $QK$ circuits in the second layer, a peculiar pattern emerges, displayed in Figures \ref{fig:L2OV} and \ref{fig:L2QK} respectively.

\vspace{-0.2 in}
The $QK$ circuit strongly resembles a \textbf{positive identity matrix}, with a distinctive blue stripe along the main diagonal, while the $OV$ circuit looks similar to a \textbf{negative identity matrix}, with a prominent red stripe along the main diagonal. This pattern holds across all second-layer heads.

Conceptually, the $OV$ circuit being a negative identity matrix implies that the source vectors added to the destination vector are simply the negative of the source vector.

Interpreting the $QK$ circuit is slightly more complex. Recall that the $QK$ circuit is a bilinear form $QK^T$ such that

\[ \vec{u} QK^T \vec{v}^T\]

determines the attention coefficient between vectors $\vec{u}$ and $\vec{v}$. If $QK^T$ were the identity, this would imply that attention between $\vec{u}$ and $\vec{v}$ is simply the dot product $\langle \vec{u}, \vec{v} \rangle$. The dot product is large when $\vec{v}$ is similar to $\vec{u}$ and small when $\vec{v}$ points in the opposite direction of $\vec{u}$. However, due to the softmax operation, the coefficient will not become negative but will instead approach zero.

Thus, we have a coherent explanation for how these circuits function together in the second layer. The $QK$ circuit identifies vectors that are similar to the destination vector, while the $OV$ circuit negatively copies these vectors. We refer to these heads as \textbf{negative similarity heads}.

\begin{figure}[H]
    \centering
    \includegraphics[width=0.75\linewidth, trim=0 0 0 10, clip]{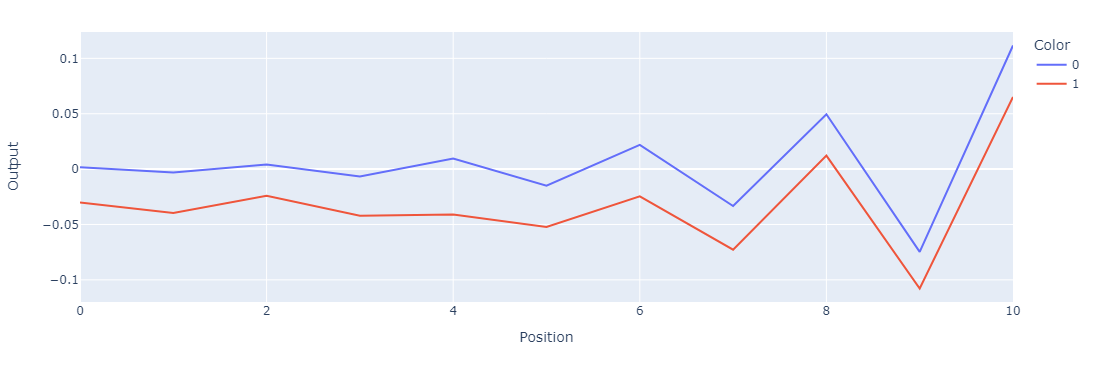}
    \caption{Residual Stream after L0 (red) vs Real Output (blue).}
    \label{fig:L0VsActual}
\end{figure}

To further validate this explanation, we can now account for the checkerboard pattern observed in the second layer attention patterns. If we project the residual stream vectors into the unembedding space directly after the zeroth layer, we observe that they already provide a reasonable estimate of the correct output, as shown in Figure \ref{fig:L0VsActual}.

Since the residual stream vectors already approximate the correct output by layer 2, the $QK$ circuit attends to every other vector starting from the destination vector itself (since these are the alternating vectors that are similar), and the $OV$ circuit negatively copies these vectors. This interaction gives rise to the distinctive checkerboard pattern in the second layer's attention patterns.

To conclude the analysis of the second layer, we formalize these notions. In the mathematical framework for transformers \cite{nanda2022glossary}, a metric known as the eigenvalue score is used to detect whether a specific head acts as a (negative) copying head.

Formally, the eigenvalue score is computed as

\[ \frac{\sum_i \lambda_i}{\sum_i |\lambda_i|}\]

where $|.|$ denotes the complex norm, and $\lambda$ denotes the eigenvalues of the $OV$ circuit. The rationale for using this metric is discussed in the cited paper.

\begin{figure}[H]
    \centering
    \includegraphics[width=0.75\linewidth]{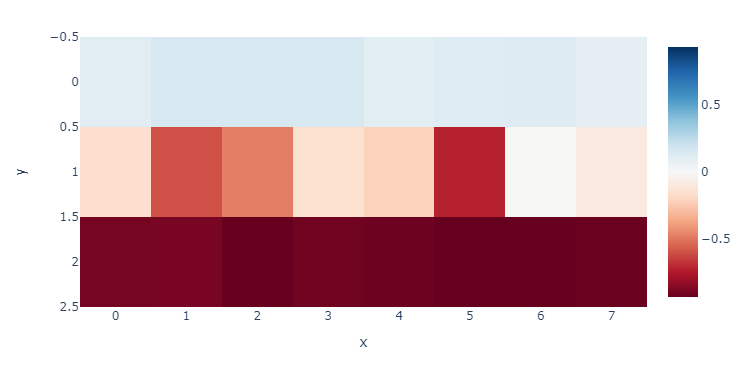}
    \caption{Eigenvalue Scores for OV Circuits}
    \label{fig:EigenValueScore}
\end{figure}

We computed this quantity for each head in each layer and displayed them in Figure \ref{fig:EigenValueScore}. We found that the second-layer heads indeed function as negative copying heads.

For the $QK$ circuit, we performed an innovative causal intervention. We aimed to replace $QK^T$ with the identity matrix; however, $QK^T$ is inherently low-rank. The best we could do was replace it with a low-rank approximation of the identity. To achieve this, we initialized $Q$ as a random matrix and set $K^T$ equal to the Moore-Penrose pseudoinverse of $Q$. After this intervention on heads 0, 1, 2, 3, and 4 in layer 2 (heads whose $QK$ circuits closely resembled the identity), we observed no degradation in accuracy (the MSE remained 0.0001). This result strongly suggests that the $QK$ circuits function as low-rank approximations of the identity matrix.

\subsection{Analysis of MLPs}

The role of multi-layer perceptrons (MLPs) in transformer interpretability remains an open problem. Recent advances, such as the use of sparse autoencoders, have made headway in this area \cite{transformer2023monosemantic}. However, since our data is strictly numerical, interpreting the different feature directions within MLPs is particularly challenging.

\begin{figure}[H]
    \centering
    \includegraphics[width=0.75\linewidth]{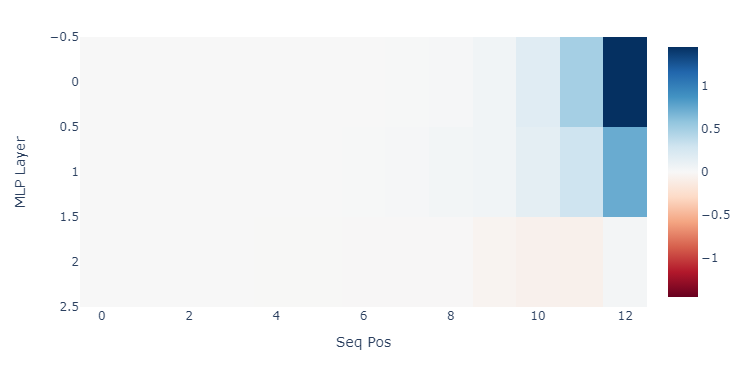}
    \caption{MLP Direct Logit Attribution}
    \label{fig:MLPLogitAttr}
\end{figure}

Nonetheless, we performed direct logit attributions on the MLPs which are shown in Figure \ref{fig:MLPLogitAttr} and found that they primarily contribute in the correct direction for the final few sequence positions.

\subsection{Conjectures on Model Functionality}

Based on the evidence presented, we propose the following conjecture regarding the model's functionality:

\begin{enumerate}
    \item The model initially forms a crude estimate of the correct solution by adding $\alpha \vec{a}_{n-1}$ to the residual stream at position $\vec{a}_n$.\\
    \item The model refines this (over)estimate using negative similarity heads, which subtract vectors that are similar to the destination vector.
\end{enumerate}

The first point is supported by our analysis of the zeroth layer's $QK/OV$ circuits, while the second point is corroborated by our analysis of the second layer's $QK/OV$ circuits.

An important question arises: why is the crude estimate $\vec{a}_n \rightarrow \vec{a}_n + \alpha \vec{a}_{n-1}$ a reasonable approximation for $\vec{a}_{n+1}$?

Substituting $\vec{a}_n$ with $c\vec{a}_{n-1} + \vec{d}$, we find that the model predicts $\vec{a}_{n+1} = (c + \alpha)\vec{a}_{n-1} + \vec{d}$. Since $\alpha > 2$ and $c + \alpha$ is strictly positive (as we bounded $c > -2$), the transformed sequence takes on a shape similar to the true output for alternating sequences.

\begin{figure}[H]
    \centering
    \includegraphics[width=0.75\linewidth]{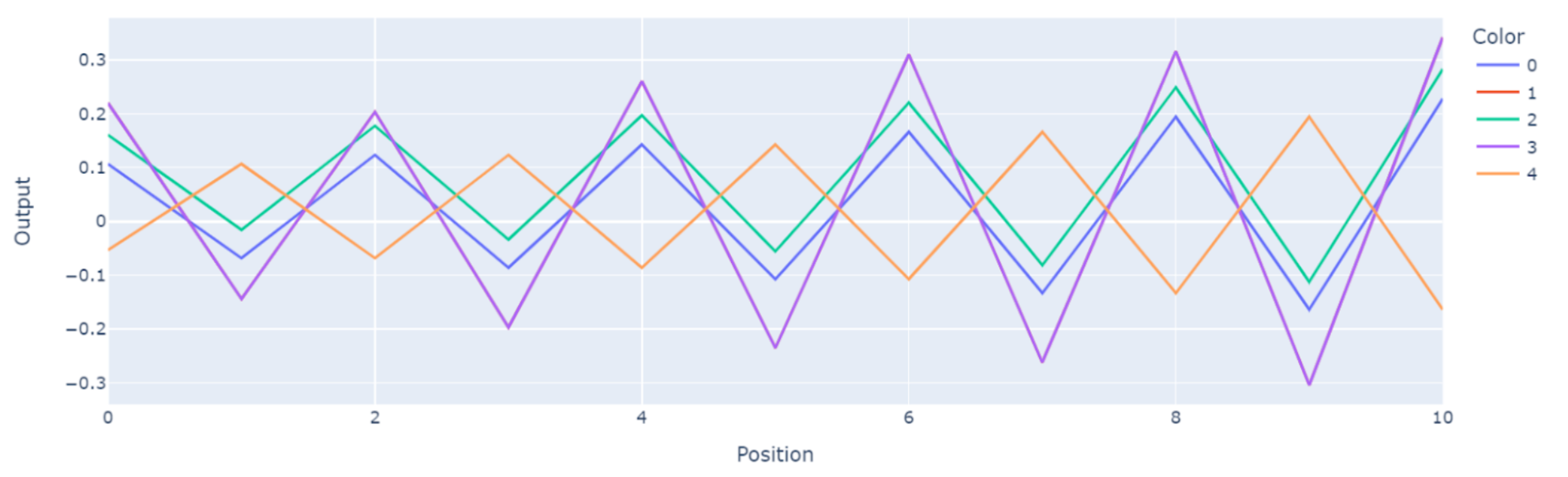}
    \caption{Residual Stream before Layer 0 (yellow), after Layer 0 (purple), with the transformation $\vec{a}_n \rightarrow \vec{a}_n + 2.3 \vec{a}_{n-1}$ (green), and actual $\vec{a}_{n+1}$ (blue).}
    \label{fig:Final}
\end{figure}

Thus, we conclude that the zeroth layer acts as a "shape-correcting" layer, providing a general approximation of $\vec{a}_{n+1}$. Subsequently, the second layer serves as a "refining" layer, correcting the initial estimate. Figure \ref{fig:Final} supports this hypothesis. 

\section{Conclusion and Future Work}

This study has provided both empirical and theoretical evidence to elucidate how GPT-style transformers solve affine recurrences. We identified that the zeroth layer leverages coefficient normalization to form a crude estimate, which is refined by negative similarity heads in the second layer.

Our primary contributions to the field of mechanistic interpretability include demonstrating that GPT-style transformers can learn affine recurrences, showcasing the utility of this task for observing transformer behaviors, and providing the weights and biases of the trained models. Additionally, we offer the code used for our analysis and a modified version of TransformerLens that can aid in the analysis of numerical tasks within transformers. We also introduced novel techniques for intervening on models, such as replacing the $QK$ circuit with the Moore-Penrose pseudoinverse to assess whether a head functions as an identity matrix and graphing residual stream vectors to detect linear correlations.

Future work could extend our analysis to higher-dimensional linear recurrences (our provided code supports this functionality), explore polynomial recurrences within transformers, and conduct a more fine-grained analysis of how transformers complete the affine recurrence task. We believe these directions could further advance our understanding of transformer behavior in numerical analysis and generalize our findings to broader contexts.

\section{Acknowledgements}

We extend our gratitude to Calum McDougall and Neel Nanda for developing the ARENA mechanistic interpretability tutorials, which provided us with much of the foundational knowledge for this study.

\vspace{-0.0in}
% {\footnotesize \bibliographystyle{ACM-Reference-Format}
% \bibliography{refs}}

\appendix

\section{Appendix}

\subsection{A Mathematical Framework for Transformers}

The conventions discussed in this section are primarily drawn from \cite{nanda2022glossary}. These conventions deviate from the standard "concatenate and multiply" perspective on transformers, offering a more interpretable framework.

A transformer, as outlined by Vaswani et al. \cite{vaswani2023attentionneed}, is a neural network architecture that maps a sequence of vectors $\vec{v}_1, \vec{v}_2, \dots, \vec{v}_N$ to another sequence of vectors $\vec{u}_1, \vec{u}_2, \dots, \vec{u}_N$.

An autoregressive decoder-only transformer is designed to predict $\vec{u}_1 = \vec{v}_2$, $\vec{u}_2 = \vec{v_3}$, $\ldots$, $\vec{u}_N = \vec{v}_{N+1}$. In other words, the model is trained to predict the next vector in a sequence. "Autoregressive" indicates that the model can only use preceding vectors in the sequence to make predictions.

Suppose the model is given $\vec{v}_1, \vec{v}_2, \ldots, \vec{v}_N$ and asked to predict $\vec{v}_{N+1}$. To achieve this, transformers must transfer information from lower sequence positions to higher ones, distributing contextual information. This process is facilitated by a mechanism known as \textbf{attention}.

Attention can be divided into two subprocesses: determining which sequence positions to transfer information from (the "source" vectors) and determining \textbf{what} information to transfer from each source vector to the destination vector. These subprocesses are embodied in the $QK$ and $OV$ circuits, respectively. The naming of these circuits will be clarified later in this section.

Essentially, the $QK$ circuit returns an \textbf{attention pattern}: a matrix where each row sums to $1$. This matrix encodes the linear combination of information from source vectors that will be added to the destination vector.

\begin{figure}[H]
    \centering
    \includegraphics[width=0.5\linewidth]{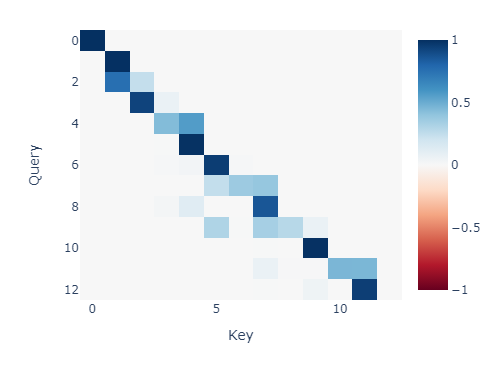}
    \caption{Attention Pattern Example}
    \label{fig:enter-label}
\end{figure}

Figure \ref{fig:enter-label} illustrates an example of an attention pattern. Each row represents the attention distribution for each destination vector. Notice that the source vector always precedes the destination vector, as the upper triangular portion of the matrix contains no attention.

Instead of directly adding linear combinations of source vectors to the destination vectors, the $OV$ circuit applies a linear transformation to each vector: $\vec{v}_1, \vec{v}_2, \ldots, \vec{v}_N \rightarrow \vec{v}_1VO, \vec{v}_2VO, \ldots, \vec{v}_NVO$. The resulting vectors are then linearly combined according to the attention pattern produced by the QK circuit and added to the corresponding destination vectors.

Each \textbf{attention head} has its own $QK$ and $OV$ circuits. Multiple attention heads operate independently, transferring information between source and destination vectors. Finally, multiple layers of these attention heads are stacked, forming the full attention mechanism.

While attention is the primary mechanism of a transformer, there are other components as well. We summarize the full transformer mechanism in the following list:

\begin{itemize}
\item The input to a transformer is a sequence of vectors $\vec{v}_1, \vec{v}_2, \dots, \vec{v}_N,$ each of which has dimension $\dv$.
\item Each vector is independently projected into a lower dimension, $\dm$, using an embedding matrix $W_E$ of dimension $\dv \times \dm$.
\item A positional embedding is added to each vector, $\vec{v}_i \rightarrow \vec{v}_i + \vec{e}_iW_P,$ where $e_i$ is the $i$'th $\text{n}_{\text{ctx}}$-dimensional one-hot vector, and $W_P$ is a $\text{n}_{\text{ctx}} \times \dm$ matrix. Here, $\text{n}_{\text{ctx}}$ is the size of the context window, or the maximum value of $N.$ 
\item The vectors are then normalized using "layer-norm," which standardizes each vector to have mean $0$ and variance $1$.
\item The attention mechanism is then applied to cross-pollinate information between different sequence positions.
\item Each vector is projected into a smaller dimension, $\dhead$, through the linear transformation $\vec{v}_i \rightarrow \vec{v}_iQ + \vec{b}_Q$, where $Q$ is a $\dm \times \dhead$ matrix and $b_Q$ is a bias term. This projection is denoted as the "query" function: $\text{query}(\vec{v}_i) = \vec{v}_iQ + \vec{b}_Q$.
\item Similarly, each vector is projected into another smaller dimension, $\dhead$, through the linear transformation $\vec{v}_i \rightarrow \vec{v}_iK + \vec{b}_K$, where $K$ is a $\dm \times \dhead$ matrix and $\vec{b}_K$ is a bias term. This transformation is denoted as the "key" function.
\item For each destination vector $\vec{v}_d$, the attention pattern is computed by first calculating

\[
\text{attn}_{s, d} = \frac{\langle \text{key}(\vec{v}_s), \text{query}(\vec{v}_d) \rangle}{\sqrt{\dhead}}
\]

for each source vector $\vec{v}_s$, where $\langle., . \rangle$ denotes the standard inner product. Note that $s \leq d$ since the transformer is autoregressive. The attention values are then softmaxed along the destination dimension to ensure that $\sum_s \text{attn}_{s, d} = 1$.
\item The $OV$ circuit determines what information is moved by applying a linear transformation $\vec{v}_i \rightarrow \vec{v}_iV + \vec{b}_V$, denoted as the "value" function.
\item For each destination token, a new vector $\vec{u}_d = \sum_{s \leq d} \text{attn}_{s, d} \text{value}(\vec{v}_s)$ is computed as the weighted sum of the value vectors, based on the attention pattern.
\item This process is repeated for each attention head, with the final $\vec{u}_d$ being the sum of the contributions from each head.
\item Each vector $\vec{u}_d$ is projected back into the $\dm$ dimension using the linear transformation $\vec{u}_d \rightarrow \vec{u}_dO + \vec{b}_O$, where $O$ is a $\dhead \times \dm$ matrix. This new $\vec{u}_d$ is added to the existing $\vec{v}_d$.
\item The vectors are normalized again using layer norm, and the output of a two-layer perceptron (MLP) is added to each vector separately: $\vec{v}_d = \vec{v_d} + \text{ReLU}(\vec{v}_dW_\text{out} + \vec{b}_{\text{in}})W_{\text{out}} + \vec{b}_{\text{out}}$. Here, $W_{\text{in}}$ is a $\dm \times \dmlp$ matrix and $W_{\text{out}}$ is a $\dmlp \times \dm$ matrix, with $\dmlp = 4\dm$ by convention.
\item This entire process is repeated for $\text{n}_{\text{layers}}$ layers, with a final layer normalization applied at the end.
\item The final vectors $\vec{v}_1, \vec{v_2}, \dots, \vec{v}_N$ are projected back into the $\dv$ space using an unembedding matrix $W_U$ of dimension $\dm \times \dv$.
\end{itemize}

One important concept is that the vector at each sequence position, $\vec{v}_i$, accumulates information from other vectors as it passes through the layers. This vector is referred to as the \textbf{residual stream vector}.

\subsection{Links to Code}
\href{https://colab.research.google.com/drive/1PEqhlGt7AIAJuIQTojTwBCca_3iqQhnh?usp=sharing}{Code to Train Transformers}

\noindent\href{https://colab.research.google.com/drive/1ZqL2pUqNJZFdIAjfuFqHCgVF4kHD3IWf?usp=sharing}{Code to Analyze Circuits}

\noindent\href{https://github.com/samarth-bhargav/recurrence_utils}{Model Repository}

\end{document}